\documentclass[letterpaper, 10 pt, conference]{ieeeconf}

\IEEEoverridecommandlockouts
\usepackage{amsmath}
\usepackage{amssymb}
\usepackage{array}
\usepackage{booktabs}
\usepackage{diagbox}
\usepackage{float}
\usepackage{graphicx}
\usepackage[hidelinks]{hyperref}
\usepackage{lipsum}
\usepackage{multirow}
\usepackage{siunitx}
\usepackage{times}
\usepackage{xcolor}
\usepackage{makecell}
\usepackage{microtype}

\usepackage{caption}
\usepackage{subcaption}

\renewcommand{\emph}[1]{\textit{#1}}
\newcommand{\revisedtext}[1]{#1}
\newcommand{\acronym}[1]{\gls{#1}\@}

\usepackage{eso-pic}
\newcommand\AtPageUpperMyright[1]{\AtPageUpperLeft{%
 \put(\LenToUnit{0.087\paperwidth},\LenToUnit{-1cm}){%
     \parbox{1.0\textwidth}{\raggedleft\fontsize{9}{11}\selectfont #1}}%
 }}%
\newcommand{\conf}[1]{%
\AddToShipoutPictureBG*{%
\AtPageUpperMyright{#1}
}
}

\usepackage[acronym]{glossaries}
\newacronym{rl}{RL}{Reinforcement Learning}
\newacronym{ros}{ROS}{Robot Operating System}
\newacronym{slam}{SLAM}{Simultaneous Localization and Mapping}
\newacronym{tsdf}{TSDF}{Truncated Signed Distance Field}
\newacronym{rms}{RMS}{Root Mean Square}
\newacronym{mdp}{MDP}{Markov Decision Process}
\newacronym{arc}{ARC}{Amazon Robotics Challenge}
\newacronym{dof}{DOF}{Degrees of Freedom}
\newacronym{ppo}{PPO}{Proximal Policy Optimization}
\newacronym{trpo}{TRPO}{Trust Region Policy Optimization}
\newacronym{tsp}{TSP}{Traveling Salesman Problem}
\newacronym{rnn}{RNN}{Recurrent Neural Network}
\newacronym{gnbv}{GNBV}{Greedy Next-Best-View}
\newacronym{ges}{GES}{Grid Exhaustive Search}
\newacronym{a3de}{A3DE}{Active 3D Exploration}
\newacronym{i3de}{I3DE}{Interactive 3D Exploration}

\DeclareRobustCommand*{\IEEEauthorrefmark}[1]{\raisebox{0pt}[0pt][0pt]{\textsuperscript{\footnotesize #1}}}

\conf{This paper has been accepted for publication at the IEEE International Conference on Robotics and Automation (ICRA), 2020. \textcopyright IEEE} 

\begin{document}


\title{\LARGE \bf Object Finding in Cluttered Scenes Using Interactive Perception}

\author{Tonci Novkovic\IEEEauthorrefmark{1}, Remi Pautrat\IEEEauthorrefmark{1}, Fadri Furrer, Michel Breyer, Roland Siegwart, and Juan Nieto%
\thanks{\IEEEauthorrefmark{1}Shared first authorship, as these authors contributed equally.}
\thanks{T. Novkovic, F. Furrer, M. Breyer, R. Siegwart, and J. Nieto are with the Autonomous Systems Lab, ETH, 8092 Zurich, Switzerland, e-mail: \{michel.breyer, fadri.furrer, tonci.novkovic\}@mavt.ethz.ch, \{rsiegwart, nietoj\}@ethz.ch.}%
\thanks{R. Pautrat is with the Computer Vision and Geometry Group, ETH, 8092 Zurich, Switzerland, e-mail: remi.pautrat@inf.ethz.ch.}%
}%

\maketitle

\pagestyle{empty}
\thispagestyle{empty}

\begin{abstract}

Object finding in clutter is a skill that requires perception of the environment and in many cases physical interaction.
In robotics, interactive perception defines a set of algorithms that leverage actions to improve the perception of the environment, and vice versa use perception to guide the next action.
Scene interactions are difficult to model, therefore, most of the current systems use predefined heuristics. 
This limits their ability to efficiently search for the target object in a complex environment. 
In order to remove heuristics and the need for explicit models of the interactions, in this work we propose a reinforcement learning based active and interactive perception system for scene exploration and object search. 
We evaluate our work both in simulated and in real-world experiments using a robotic manipulator equipped with an RGB and a depth camera, and compare our system to two baselines.
The results indicate that our approach, trained in simulation only, transfers smoothly to reality and can solve the object finding task efficiently and with more than $\bf 88\%$ success rate.

\end{abstract}



\section{Introduction}

Often robotic systems do not rely on physical interactions to improve their perception of the world.
One application where such interactions could significantly improve the performance is object search in unstructured environments.
By doing so, robots would be able to uncover previously unobservable parts of the environment and therefore have a more complete interpretation of the world.

Some methods from computer vision focus on detecting objects in single RGB images.
With recent progress in deep learning, such methods have shown impressive results, even surpassing human performance in the object detection task~\cite{he2017mask, redmon2017yolo9000, li2017}.
However, a single image is usually not sufficient to find an object if it is hidden in clutter and occluded by other objects.
Because of this, \emph{active perception}, where the camera is actively moved, is required to observe the scene from different viewpoints.
In 3D scene reconstruction~\cite{kriegel2015efficient}, changing the camera viewpoint is required to reveal details of the scene that are obstructed by other objects~\cite{kahn2015active} or to gain more information on the scene in order to grasp an object~\cite{morrison2019multiview}.
Selecting the ``next best view'' where the camera should move can be based on the current knowledge of the scene and the given task~\cite{Dunn2009NextBV}, or selected based on a greedy heuristic~\cite{delmerico2018comparison} that does not account for long term planning reward.

Even by allowing camera motions and smart selections of the next views, objects can still be hidden in a pile or occluded by other objects.
Therefore, interaction with the environment is necessary to remove these occluders.
\emph{Interactive perception} builds upon \emph{active perception} and combines camera motion with environment interaction.
It uses information from interactions to get a better understanding of the scene. 
Such systems have been demonstrated for different applications such as segmentation, grasp planning or object recognition~\cite{bohg2017interactive}, even using \acronym{rl}~\cite{cheng2018reinforcement}. 
\revisedtext{However, they usually use predefined actions~\cite{danielczuk2019mechanical} and do not consider exploration by assuming that the target is always within the view of the camera.}
As a result, such methods are constraining the robot to a limited set of interactions which are usually not efficient for object search in cluttered environments.

\begin{figure}[t!]
  \centering
  \includegraphics[width=\columnwidth]{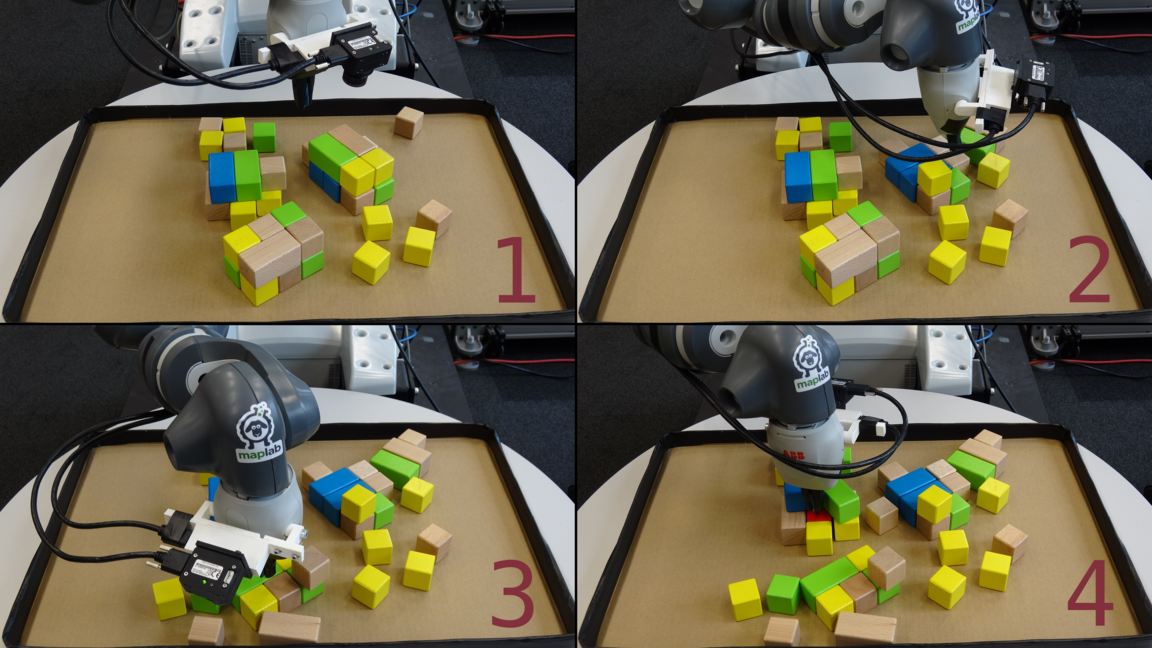}
  \caption{In order to find a specific object in a scene, it is sometimes necessary to interact with the environment. In our experiments, the robot learned that the object of interest, i.e. the red cube, can be hidden inside a pile of cubes and that physically interacting with the objects might reveal it.}
  \label{fig:teaser_image}
\end{figure}

Our approach to object finding in clutter, using interactive perception, is built upon an \acronym{rl}-based control algorithm and a color detector.
Since object detection in a single image is already a very well studied problem and to simplify our approach, we assume that the target object is of a specific color.
In addition, we encode the scene state using a discretized \acronym{tsdf} volumetric representation~\cite{curless1996tsdf}.
The agent's next action is determined by an \acronym{rl} algorithm based on the current encoded state and the knowledge obtained from past experiences.
Since specifying which actions are good and which are bad is very hard and non-intuitive, supervised methods are not adapted to this task.
With such a framework, we show that we can learn a policy that can effectively search an object in a cluttered scene.
The main contributions of this paper are:
\begin{itemize}
    \item an \acronym{rl} approach to \emph{active} and \emph{interactive perception} based object search in clutter,
    \item a compressed volumetric representation of the environment suitable for \acronym{rl}-based object finding,
    \item experimental evaluation of the framework on both a simulated and a real-world robotic system, including a comparison to baselines.
\end{itemize}

\section{Related Work}\label{sec:related_work}

Finding objects hidden in clutter requires robots to actively explore and manipulate their environment.
In this section, we present an overview of previous works related to active and interactive perception, and their application to manipulation.


In contrast to recent works in object detection and semantic segmentation that typically only operate on single fixed images~\cite{he2017mask,redmon2017yolo9000}, active perception considers the problem of optimizing the sensor placement in order to perform a task~\cite{bajcsy2018revisiting}, and has found applications in a variety of tasks such as mapping~\cite{carrillo2012comparison} and object reconstruction~\cite{pito1999solution, wenhardt2007active, kriegel2015efficient}.
A common approach is to choose the next-best-view in order to maximize an information gain metric~\cite{connolly1985determination}.
Velez et al.~\cite{velez2011planning} presented a planner for improved object detection in an unknown scene taking into account the confidence of the object detector and uncertainty of the robot's pose.
Choosing a good metric is often challenging and task specific~\cite{chen2011active}, however measures such as the Shannon entropy~\cite{vazquez2001viewpoint} or the KL-divergence~\cite{hoof2012maximally} are often used.
Recently, new information gain formulations have been proposed for volumetric reconstruction of unknown scenes~\cite{isler2016information, delmerico2018comparison}.
Acquiring more information of a scene is also beneficial for manipulation.
Kahn et al.~\cite{kahn2015active} model the probability of grasp locations in occluded regions as a mixture of Gaussians and optimize for sensor placements that minimize uncertainty of these regions, while Morrison et al.~\cite{morrison2019multiview} choose informative viewpoints based on a distribution of grasp pose estimates.


However, in many cases robots can improve perception through physical interaction with their environment.
The goal of interactive perception is to learn the relationship between actions and their sensory response~\cite{bohg2017interactive}.
Applications cover a wide range of problems, such as object recognition~\cite{sinapov2014learning}, object segmentation~\cite{fitzpatrick2002manipulationdriven, hoof2012maximally}, and inferring physical properties of objects~\cite{xu2019densephysnet, martin-martin2017building}.
Dogar et al.~\cite{dogar2014object} generate plans that minimize the expected time to find an occluded object using a visibility-accessibility graph, assuming that the target is the only hidden object in the scene.
Li et al.~\cite{li2016act} formulate object search as a POMDP and solve it with an approximate online solver.
Their approach however relies on highly accurate segmentations.
Xiao et al.~\cite{xiao2019online} address some of these limitations, assuming knowledge of the exact number and the geometric properties of the objects in the scene.
In a recent work, Danielczuk et al. use the output of a grasp planner~\cite{mahler2019learning} and different heuristics to choose between grasping, suction, and pushing actions to extract a target object from a heap~\cite{danielczuk2019mechanical}. 


The majority of these works rely on predefined action primitives, such as pushing and grasping, reducing the set of possible actions to choose from.
In contrast, we propose to use \acronym{rl} in order to learn a suitable policy and allow the agent to freely control the pose of its end-effector.
Gualtieri et al.~\cite{gualtieri2018} also use \acronym{rl} to learn 6 \acronym{dof} movement of a robot arm.
However, to simplify the problem, they constrain the agent to focus attention on sub-regions of the current observation, which requires the target to be visible from the beginning.
In our previous work~\cite{breyer2019comparing}, we leverage \acronym{rl} to design simple actions in order to clear a table of objects, with a camera mounted on the end-effector.
Cheng et al. \cite{cheng2018reinforcement} train a \acronym{rnn} to predict gripper displacements that push an object to a target location, while being robust to occlusions.


A critical component of the latter work was to incorporate the prediction of an object detector in the observation encoding.
This agrees with the findings of Sax et al.~\cite{sax2018midlevel}, who motivate that task-specific, mid-level features offer better learning efficiency and generalization compared to raw images.
This work thus tries to focus on simple and interpretable inputs, allowing efficient learning of interactive tasks.

\section{Method}\label{sec:method}

\begin{figure}
  \centering
  \includegraphics[width=0.95\columnwidth]{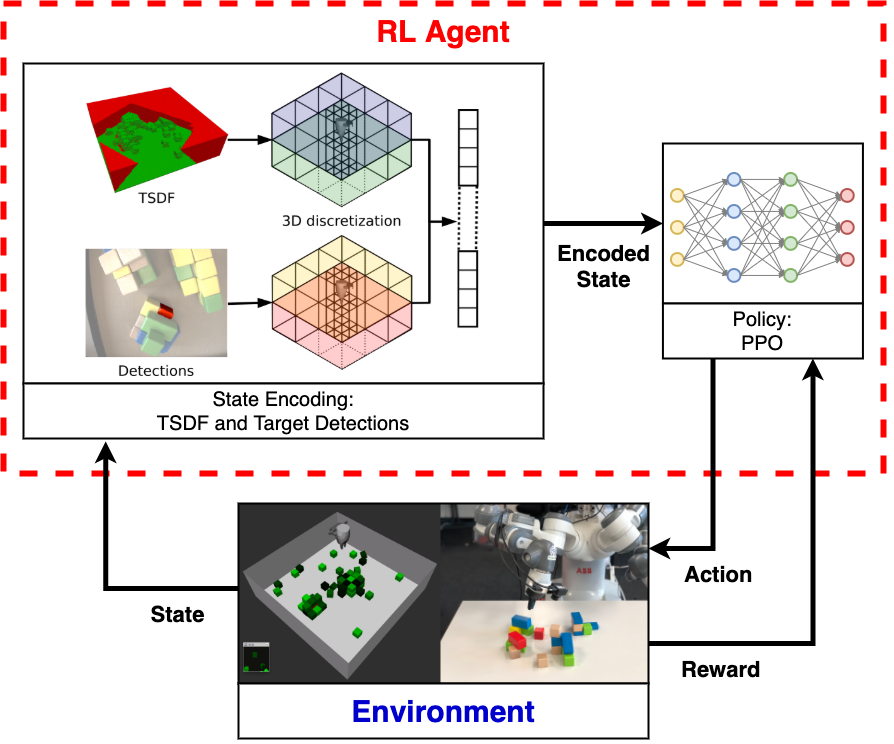}
  \caption{To find the objects in clutter, we encode the sensor measurements from the RGB and the depth camera into a vector that is then given to the \acronym{rl} agent. The vector is generated from the volumetric \acronym{tsdf} map and the object detections around the gripper. This information is summarized by discretizing the space into bins centered in the end-effector frame and projected onto the $xy$-plane. The final state vector is then composed of these 68 values, 2 normalization factors for the \acronym{tsdf} and detection part and the current tilt angle value. The agent computes the next action which is executed by the robot and a feedback, a reward, is provided to the agent. Once the agent finds the object, claims that there is no object in the scene, or reaches the maximum number of time steps, the execution is terminated. 
  }
  \label{fig:system_overview}
\end{figure}

We tackle the problem of finding specific objects in cluttered environments where interaction is required, using a robotic arm with wrist-mounted RGB and depth cameras.
Our goal is to find a mapping from sensor inputs to end-effector displacements, i.e. a policy $\pi$, using a model-free \acronym{rl} algorithm. Figure~\ref{fig:system_overview} shows the system overview.

\subsection{Problem Formulation} \label{sec:problem_formulation}

\revisedtext{
In our problem formulation, the state is only partially observable.
However, we model it as a discrete, finite horizon \acronym{mdp} such that we can rely on standard \acronym{rl}-based methods to solve it.
}
We denote the set of all states as $S$; $A(s)$ is the set of all valid actions for a given state $s$, $r : S \times A \times S \to R \subset \mathbb{R}$ a real-valued reward function, $\gamma$ the discount factor, and $p : S \times R \times S \times A \to [0,1]$ a deterministic function that defines the dynamics of the \acronym{mdp}.

At each time step $t$, the \acronym{rl} agent observes the current state $s_t \in S$.
Based on the parameterized policy $\pi_\theta(a_t | s_t)$, it decides which action $a_t \in A$ to take next.
Upon completion of this action the system transitions into a new state $s_{t+1} \in S$ based on the system dynamics $p$ and receives a reward $r_{t+1}(s_t, a_t, s_{t+1})$.
For each episode (one run of the algorithm) this is repeated until the maximum number of time steps $T$ is reached or the agent reaches the terminal state.
The final goal of the on-policy \acronym{rl} is to find the optimal actions for the given states that maximize the expected return, i.e. the total discounted reward.

\subsection{Agent Model} \label{sec:agent_model}

Our algorithm first converts raw sensor measurements into a simplified \acronym{rl} state representation.
The agent, based on the current state, plans the next action for the robot.
In order to plan meaningful actions, the agent is trained in simulation where a reward is given for each episode.

\subsubsection{State Representation}
In our object finding task, the representation has to allow the agent to reason about the spatial relationships between the objects, to distinguish the target object, and to provide information about unobserved areas.
However, the representation should also be compact enough to keep the network size reasonable and informative enough to allow the convergence of the \acronym{rl} agent.


We use a volumetric \acronym{tsdf}~\cite{curless1996tsdf} representation to provide us with spatial information about the scene.
However, since we are interested in specific objects, we have extended this voxel grid with an additional voxel value that indicates if the voxel is part of the target object or not.
\revisedtext{The target is first detected in the RGB image by an object detector.
These detections are then projected to 3D using the depth information from the depth image and integrated into a truncated \emph{unsigned} distance field.
Similarly as for the \acronym{tsdf}, we use voxel carving to remove any detection in free space.}
Since the focus of our work is on interactive perception, we simplify the detection problem by using a target with a specific color, so that a color detector in CIELAB color space can be used to detect it.
\revisedtext{However, this does not limit our approach since detections obtained by any other detector can be integrated into the proposed volumetric representation in an identical manner.}

Depending on the resolution of the voxels, such a representation can contain a very large amount of information.
Since it would be infeasible to use this information directly in an \acronym{rl} algorithm, we encode it into a smaller sized vector that can be efficiently used for learning.
\revisedtext{To do this, we first transform the \acronym{tsdf} enriched with the detection channel to the end-effector frame projected onto the $xy$-plane.
Thus, the grids are centered on the end-effector's position in the $xy$-plane and are rotated around the $z$ axis such that their $x$ and $y$ axis are aligned with the projected end-effector's $x$ and $y$ axis.
We also crop the two volumetric representations in a square centered on the robot to be able to scale to large environments.
Thus, the 3D representation contains only the local \acronym{tsdf} and detection values around the robot, whereas the scene and the \acronym{tsdf} can be much bigger.
The two 3D grids are then split into two layers, above and below the fingers of the gripper.
Each layer is summed along the $z$ coordinate to get $4$ flattened maps.
These maps are then discretized into a $3 \times 3$ grid, where the middle cell is further separated into a $3 \times 3$ grid.
Finally, the $2$ layers coming from the \acronym{tsdf} and the $2$ other layers containing the detections are normalized separately, respectively by the sum of the \acronym{tsdf} map values and the sum of the detection map values.
As a result, we obtain $4$ maps encoded with $17$ values, which are then concatenated into a $68$ dimensional vector, as depicted in Figure~\ref{fig:system_overview}.
The final state vector is composed of $71$ values: the previously defined $68$ dimensional vector, the $2$ normalization factors (to keep track of how well the scene has been explored), and a scalar indicating the current tilt angle of the gripper.
Even though the 3D maps are defined in a frame relative to the robot, because we use a projected frame, the tilt of the robot cannot be observed, therefore it is added to the state.
}


To handle modified environments after interactions, we limit the maximum weight in the \acronym{tsdf} grid to $2$. 
This means that all past measurements have a small weight (high uncertainty) and when new measurements are obtained, old ones are quickly forgotten.
Such a strategy allows us to update the representation very quickly by only taking a few additional measurements. 


\subsubsection{Actions}
To perform the object finding task successfully, the robot is required to move the camera in 3D, as well as to interact with the objects in the scene.
To generalize well to both tasks of active and interactive perception, we decided to control relative displacements of the gripper in the end-effector frame.
\revisedtext{More precisely, all the actions are performed in a frame centered at the end-effector and rotated around the $z$-axis by the yaw of the gripper compared to the global frame.
We have 3 variables for the translations in $x$, $y$, and $z$ direction.} 
Since exploring the unobserved areas can be much faster by tilting the end-effector, we additionally added roll ($\phi$) and yaw ($\psi$) angles.
A binary variable allows the agent to terminate the episode to express that it either found the target object or to indicate that no target object is present.

\subsubsection{Reward Function}
To improve the convergence of \acronym{rl} and avoid local optima, we use the following rewards:
\begin{itemize}
    \item Time penalty - to encourage the agent to finish the task quickly, a negative reward $r_t$, is given at each step.
    \item Exploration reward - the agent is rewarded at each step for exploring the scene, proportionally to the number of newly observed voxels.
    \revisedtext{This reward is always smaller than or equal to the negative time penalty, $r_t$.}
    \item Final detection reward - the agent gets a final detection reward $r_d$ if it terminates after observing the target.
    \item Final exploration reward - if there is no target in the scene and the agent terminates, it gets a final exploration reward equal to the final detection reward $r_d$.
    \item Final failure penalty - if the agent terminates before finding the object in the scene, or reaches a maximum number of steps in an episode, it gets a failure penalty equal to the negative final detection reward $-r_d$.
\end{itemize}
At each iteration, if the agent does not terminate or find the object, the best that it can achieve is a $0$ reward since the maximum exploration reward is equal to the time penalty. 
Getting a total positive reward means that the agent decided to terminate after having seen the target or when there was no target in the scene.

\subsubsection{Policy} \label{sec:policy}
To find a successful policy, we used a fully connected network to map the current state to the next action and update its weights with \acronym{ppo}~\cite{SchulmanWDRK17}, a state-of-the-art policy gradient method.

\subsection{Simulation} \label{sec:simulation}
In order to train the \acronym{rl} model, we use a simulation environment based on the Bullet physics engine~\cite{Coumans2015Bullet}.
\revisedtext{The simulated scene is surrounded by walls, and initial positions and number of objects are randomized in each episode.
We generate five different types of scenes that contain: (i) only \emph{cubes}, (ii) \emph{fixed size primitives} (cubes, cuboids, cylinders and prisms), (iii) \emph{variable size primitives} (from $0.5$ to $2$ times the original size), (iv) \emph{random shapes}\footnote{https://sites.google.com/site/brainrobotdata/home/models}, and (v) realistic household \emph{object models}.}
The robot is simulated with a gripper without an arm whose pose is controlled by a force constraint.
Furthermore, we attached a virtual RGB-D camera to the end-effector, mimicking the real robot setup. 
Depth and RGB images are generated using bullet's camera renderer.
\revisedtext{To simulate a realistic object detector with uncertain detections, we introduce false detections inside our color detector and add an uncertainty measure for each detected pixel.
We subtract from each true detection (with an original value of 1) a random sample from an exponential distribution with parameter $\lambda = 20$ to model the fact that detections are uncertain.
To model false positives, $2\%$ of the pixels in the image are randomly selected and assigned a value from another exponential distribution with parameter $\lambda = 20$.
}

\subsection{Transfer to Reality} \label{sec:transfer_to_reality}
Transfer to reality was done without any fine-tuning of the model on the real robot.
Since the robot poses are provided in the base frame of the robot, and the depth measurements are in the camera frame, the transformation between the two frames is required to properly integrate the measurements into the volumetric grids.
The transformation was obtained using our hand-eye-calibration toolbox~\cite{furrer2018evaluation}.
The fact that we use a distance field representation, which averages depth images during integration time, allows us to have some noise in the depth image and small errors in the pose estimates of the camera. 
The discretization of the volume used to represent the state, additionally, allows small scene reconstruction errors. 
Such a choice of state makes the precise details of the scene less important, and as a result, allows for a better abstraction and generalization of real data.

\section{Evaluation}\label{sec:evaluation}

In our experiments, we evaluate the performance of our approach both in simulation and in the real world, and compare it to two baselines, one for the active perception and one for the interactive perception task.
Additionally, we assess how the approach generalizes to different scenes.
As a metric for our evaluation we use success rate, number of steps per episode, and total time per episode.

\subsection{Experimental Setup}

\begin{figure}
  \centering
    \begin{subfigure}[]{0.492\columnwidth}
        \includegraphics[width=1.0\columnwidth]{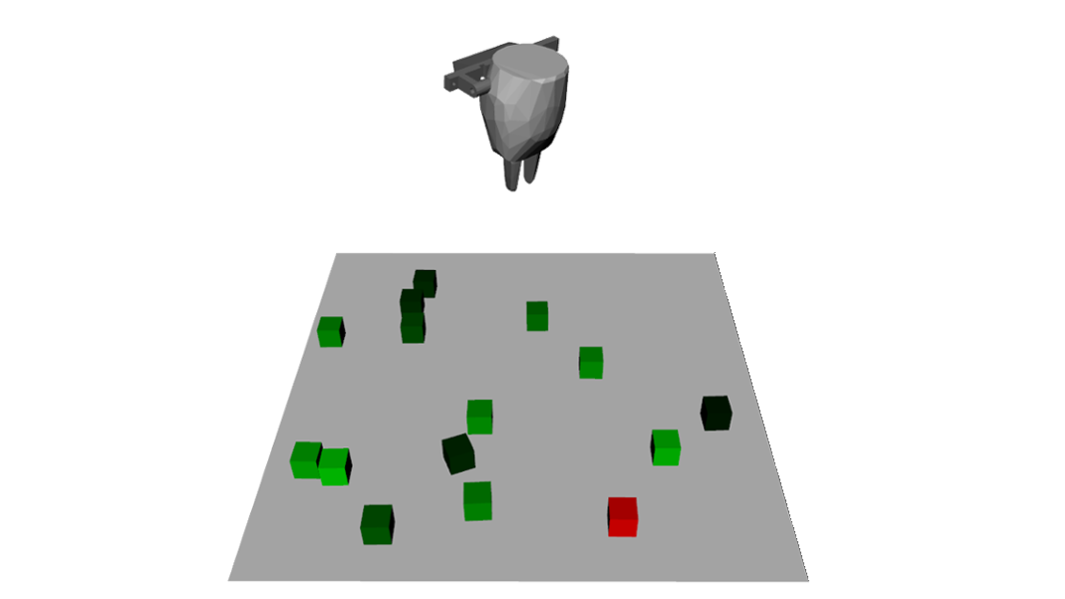}
        \caption{AP simulation} 
        \label{fig:setup_ap_simulation}
    \end{subfigure}
    \begin{subfigure}[]{0.492\columnwidth}
        \includegraphics[width=1.0\columnwidth]{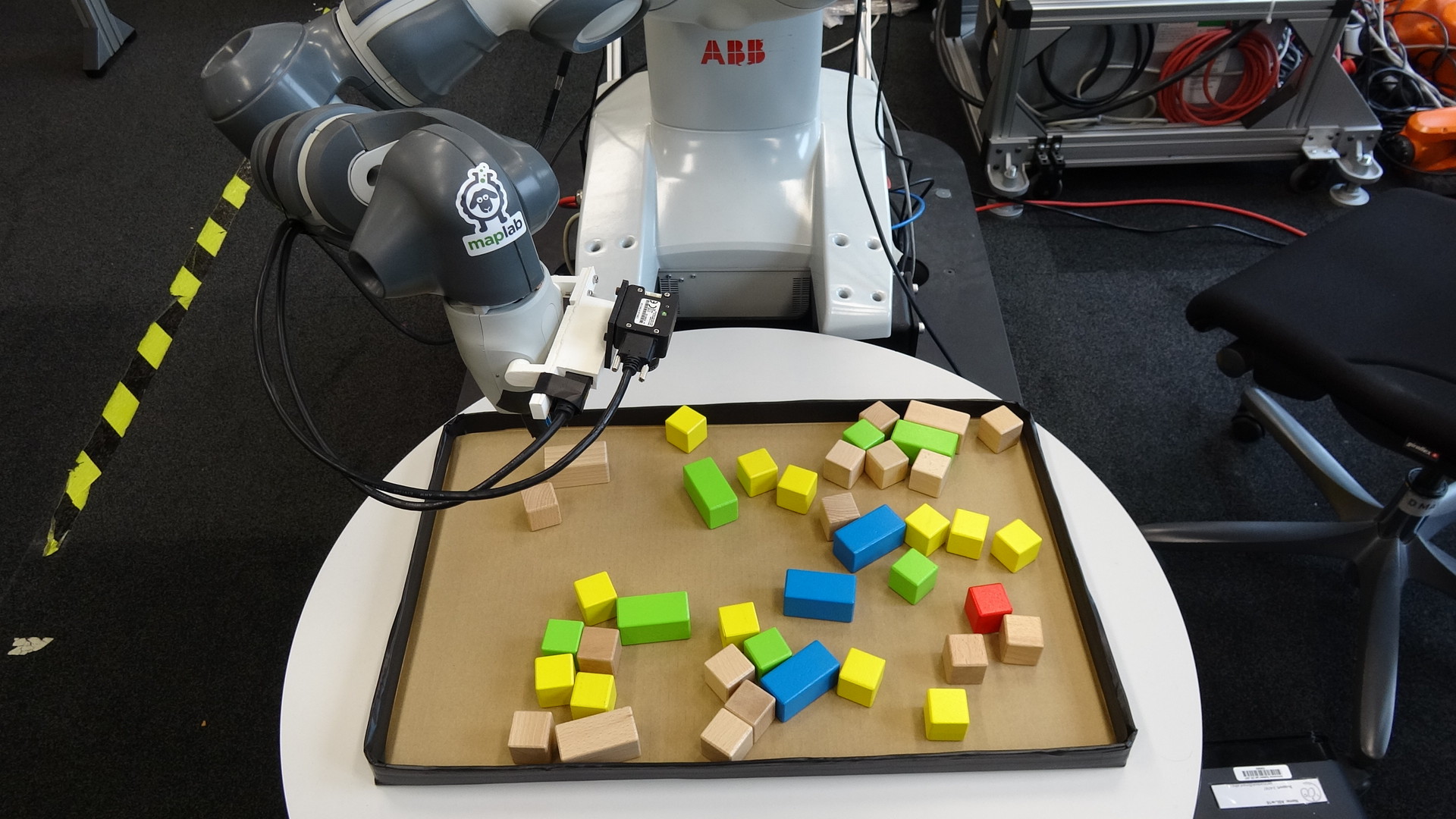}
        \caption{AP real world} 
        \label{fig:setup_ap_real_world}
    \end{subfigure}\\
    \begin{subfigure}[]{0.492\columnwidth}
        \includegraphics[width=1.0\columnwidth]{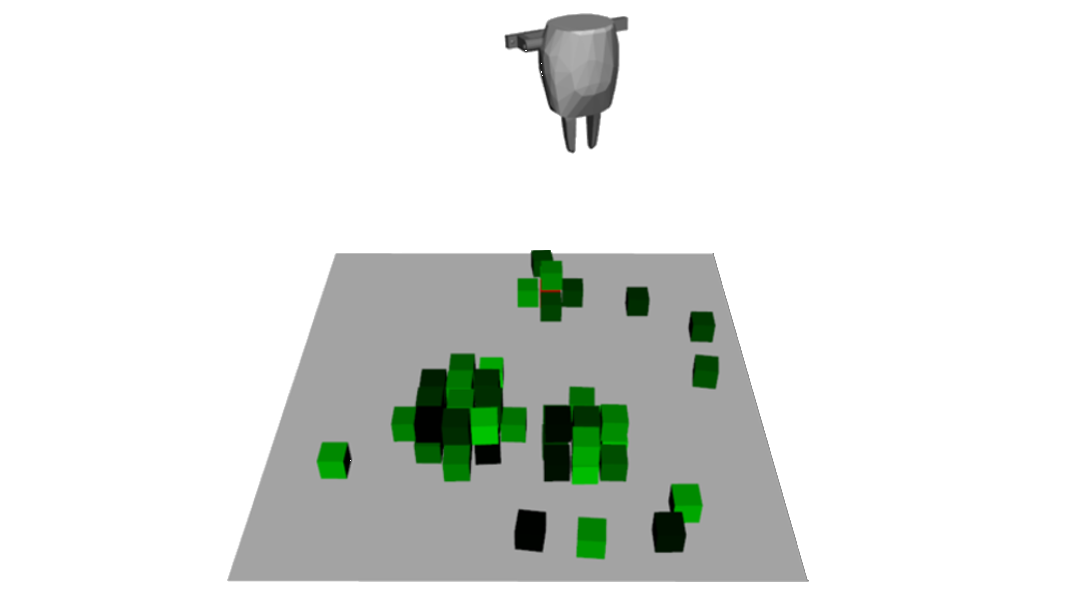}
        \caption{IP simulation} 
        \label{fig:setup_ip_simulation}
    \end{subfigure}
    \begin{subfigure}[]{0.492\columnwidth}
        \includegraphics[width=1.0\columnwidth]{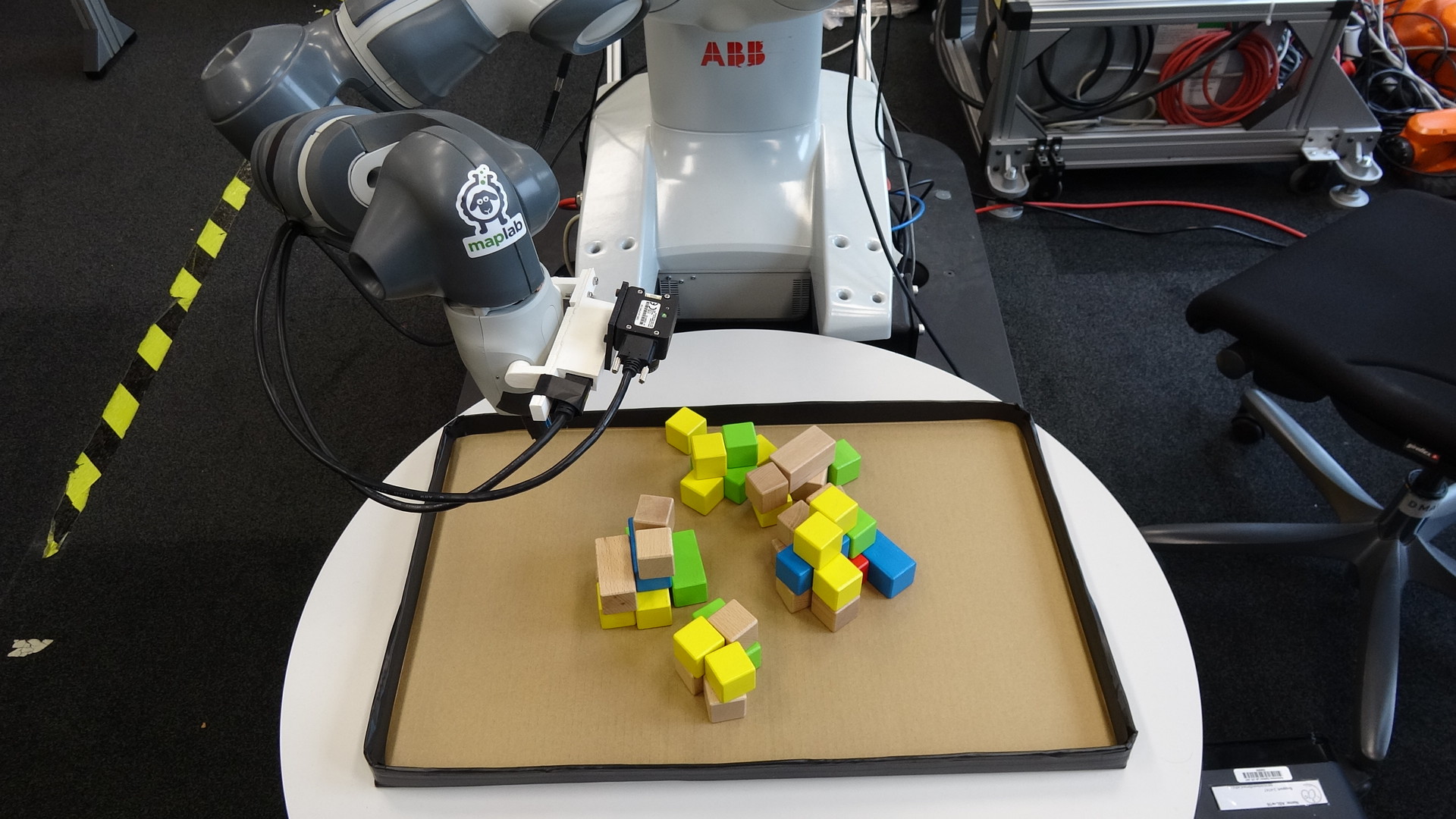}
        \caption{IP real world} 
        \label{fig:setup_ip_real_world}
    \end{subfigure}
  \caption{Setup of the simulation and real-world experiments for the active and interactive perception tasks of object finding. For the active perception (AP) task, the target object is never completely covered by the other objects and can be found by moving the camera. In the interactive perception (IP) task, the object can be hidden within a pile of objects and, therefore, interaction is necessary to reveal it. The hand can freely move within a defined workspace in simulation, whereas it has to obey kinematic rules in the real world.}
  \label{fig:setup_sim_and_real}
\end{figure}

We evaluate our method within a simulated environment, as described in~\ref{sec:simulation}, and in the real world, using a position controlled 7-DoF arm of an ABB Yumi. 
An RGB Chameleon~3 camera, coupled with a CamBoard pico flexx depth camera, is mounted on the wrist of the robot. 
We use a similar model in the physics engine Pybullet~\cite{coumans2018} to perform the task realistically in simulation. 
The position of the gripper is randomly initialized.

The objects in the scene are either scattered on the ground in the active task or forming piles in the interactive one. 
The target, when there is one, is always the single red object in the scene.
The different setups with scenes containing \emph{cubes} are visualized in Figure~\ref{fig:setup_sim_and_real}.
We used Open3D~\cite{Zhou2018} and modified its \acronym{tsdf} representation to integrate distance values together with target detections.
The target detection is performed by first converting the RGB image into CIELAB color space.
Red pixels are labelled as belonging to the target.
Finally, false detections and the detection score are added to each~pixel.

Both tasks are trained in simulation using the \acronym{ppo} implementation of the OpenAI Baselines framework~\cite{baselines} with a batch size of $5000$, a minibatch size of $500$, a discount factor of $0.99$, an entropy bonus of $0.01$, and reward values $r_t=1$ and $r_d=150$.
The network used for inference is a Multi Layer Perceptron with $2$ hidden layers of size $200$.

We set the \acronym{tsdf} truncation distance to $0.04$\,\si{m} and represent the workspace by a volumetric grid with a resolution of $100$ voxels per side.
The state representation is cropped by a $1.2$\,\si{m} square around the end-effector center.
The maximum translation per step of the agent is $0.06$\,\si{m} and $0.15$\,\si{rad} for the roll and yaw rotation.


\begin{table}
    \centering
    \begin{tabular}{@{}rccccc@{}}
    \toprule
     & \multicolumn{3}{c}{Simulation} & \multicolumn{2}{c}{Real World} \\
     \cmidrule(lr){2-4}  \cmidrule(lr){5-6}
     & AP & AP (L) & IP & AP & IP \\ \midrule
     Min num. objects & $5$ & $10$ & $15$ & $15$ & $15$ \\
     Max num. objects & $25$ & $40$ & $75$ & $30$ & $30$ \\
     Num. of piles & $0$ & $0$ & $3$ & $0$ & $4$ \\
     No target prob. & $0.10$ & $0.10$ & $0.10$ & $0.0$ & $0.0$ \\
     Exploration ratio & / & / & $0.25$ & / & $0.1$ \\
     Workspace length [m] & $0.6$ & $1.2$ & $0.6$ & $0.33$ & $0.33$ \\
     Time horizon & $130$ & $500$ & $130$ & $50$ & $50$ \\
    \bottomrule 
    \end{tabular}
    \caption{Simulation and real-world experiment parameters for both normal and large (L) scenes in active perception (AP), and interactive perception (IP) tasks. Exploration ratio specifies a percentage of scenes for which there are no piles, therefore it is only used in the interactive task.}
    \label{tab:experiment_parameters}
\end{table}

\subsection{Baselines}
In order to evaluate the performance of our algorithm, we have implemented two baselines, one for the active perception task and one for the interactive perception task.

\subsubsection{Active Perception}
For the active perception task, where the target object can be revealed just by moving the camera, we have implemented a one-step \acronym{gnbv} agent~\cite{connolly1985determination}.
This agent samples $n$ poses within the proximity of the current pose and evaluates each of them based on how many voxels would be observed if the agent would have moved to that location, assuming that there are no objects in the unobserved area.
The sample that obtains the highest score is used to determine the next action.
\revisedtext{Note that the greedy agent has access only to the local \acronym{tsdf} values around the gripper, similarly to the the \acronym{rl} agent, to allow a fair comparison.}
If one part of the red object is detected, then the samples are evaluated based on which one brings the target object closer to the camera image center.
Once the agent explores more than $97\%$ of the voxels (the agent decides that there is no target in the scene), or it sees more than $20\%$ of the red object, it terminates the search.
This means that in most cases the scenes without a target object are successfully completed after enough steps.
In our experiments, we used $n=10$ pose samples in simulation and $n=30$ pose samples in the real-world experiments.
This is due to the fact that in the real world, finding a suitable sample is harder since kinematic constraints of the robotic arm need to be considered, which is not the case in simulation.

\subsubsection{Interactive Perception}
\revisedtext{Comparing the interactive task to a baseline is not straightforward.
To the best of our knowledge, there is no method that is able to both explore and interact with the environment for object search. 
}
We instead compare our approach with a \acronym{ges} agent.
The agent traverses the whole workspace at a fixed low height (just above the lowest cube) pushing any cube along the way that is placed on top of another cube, thus revealing any hidden cubes below.
To achieve this, the agent divides the workspace into a regular grid which defines a graph whose nodes are intersections of the grid.
The nodes are one maximum translation allowed apart from each other such that the agent can move from one node to another in one step.
Since the initial position of the agent is randomized, the problem is equivalent to the \acronym{tsp} and is NP-complete.
Therefore, we used the 2-opt algorithm~\cite{Croes58} to efficiently compute a route that is close to optimal.
The robot then traverses the route until it has seen more than $20$\% of the object or goes through the whole grid, in which case it decides that there is no target in the scene.
This indicates that all the scenes without the target object will be successfully completed unless the agent decides to terminate due to false detections.

\begin{figure}
  \centering
    \begin{subfigure}[]{0.32\columnwidth}
        \includegraphics[width=\columnwidth]{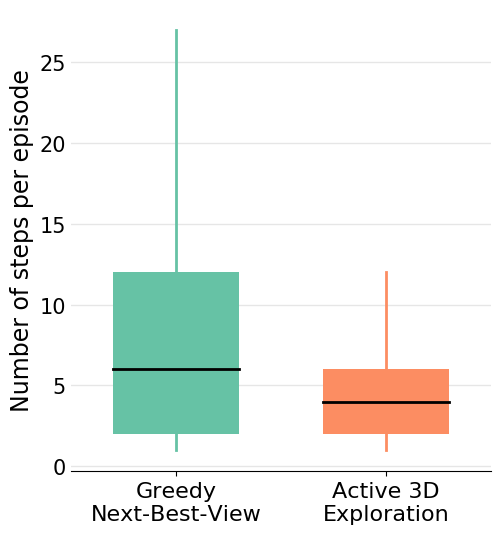}
        \caption{} 
        \label{fig:simulated_results_efficiency_active_small}
    \end{subfigure}
    \begin{subfigure}[]{0.32\columnwidth}
        \includegraphics[width=\columnwidth]{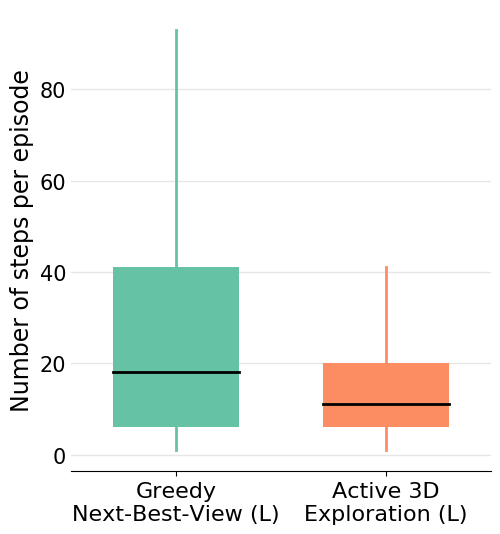}
        \caption{} 
        \label{fig:simulated_results_efficiency_active_medium}
    \end{subfigure}
    \begin{subfigure}[]{0.32\columnwidth}
        \includegraphics[width=\columnwidth]{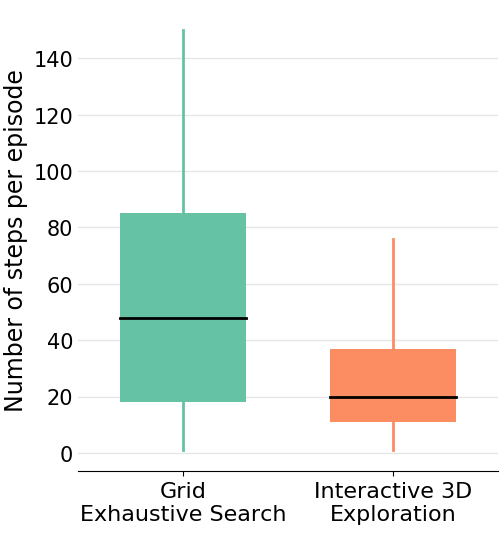}
        \caption{} 
        \label{fig:simulated_results_efficiency_interactive}
    \end{subfigure}
  \caption{Simulation results: Average number of steps per episode among the active methods on \emph{cubes} scenes (\subref{fig:simulated_results_efficiency_active_small}), larger \emph{cubes} scenes (\subref{fig:simulated_results_efficiency_active_medium}) and interactive methods on \emph{cubes} scenes (\subref{fig:simulated_results_efficiency_interactive}).}
  \label{fig:simulated_results_efficiency}
\end{figure}

\subsection{Simulation Results}

\begin{table*}[t]
    \centering
    \resizebox{\textwidth}{!}{%
    \begin{tabular}{ccccccccccccccc} 
    & \multicolumn{4}{c}{\includegraphics[width=0.12\textwidth]{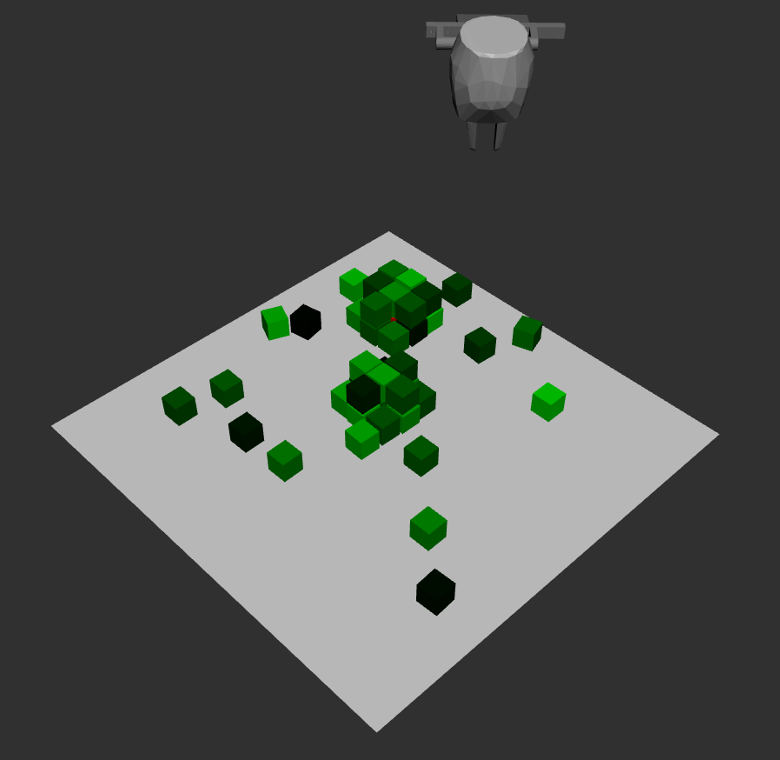}} & \multicolumn{2}{c}{\includegraphics[width=0.12\textwidth]{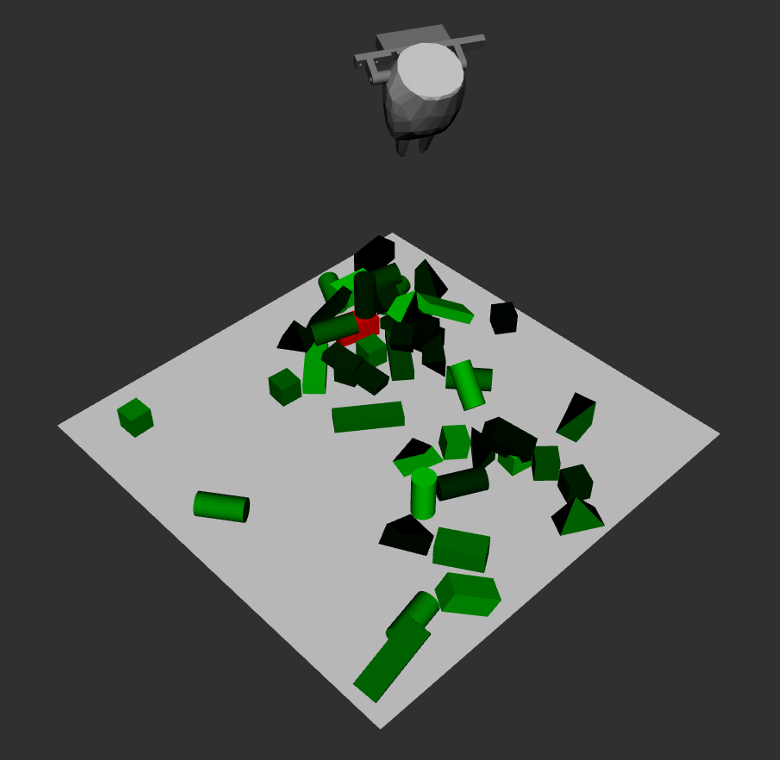}}
    & \multicolumn{2}{c}{\includegraphics[width=0.12\textwidth]{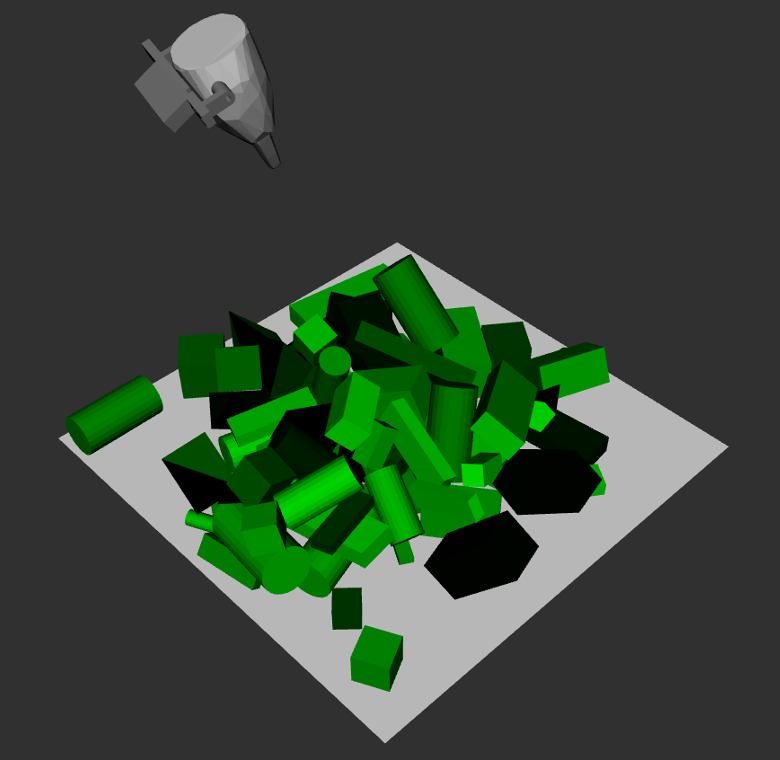}} &
    \multicolumn{2}{c}{\includegraphics[width=0.12\textwidth]{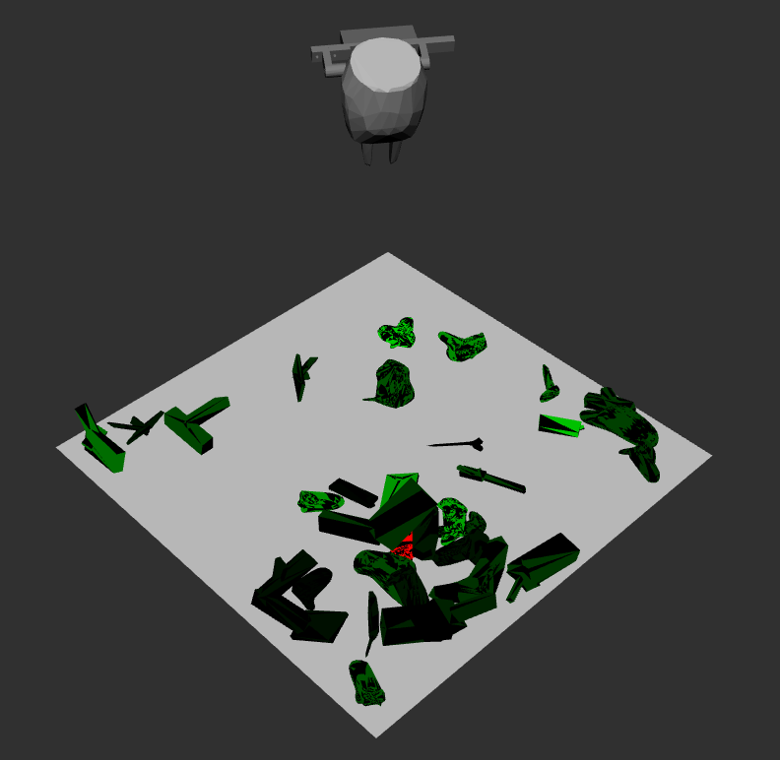}} &
    \multicolumn{2}{c}{\includegraphics[width=0.12\textwidth]{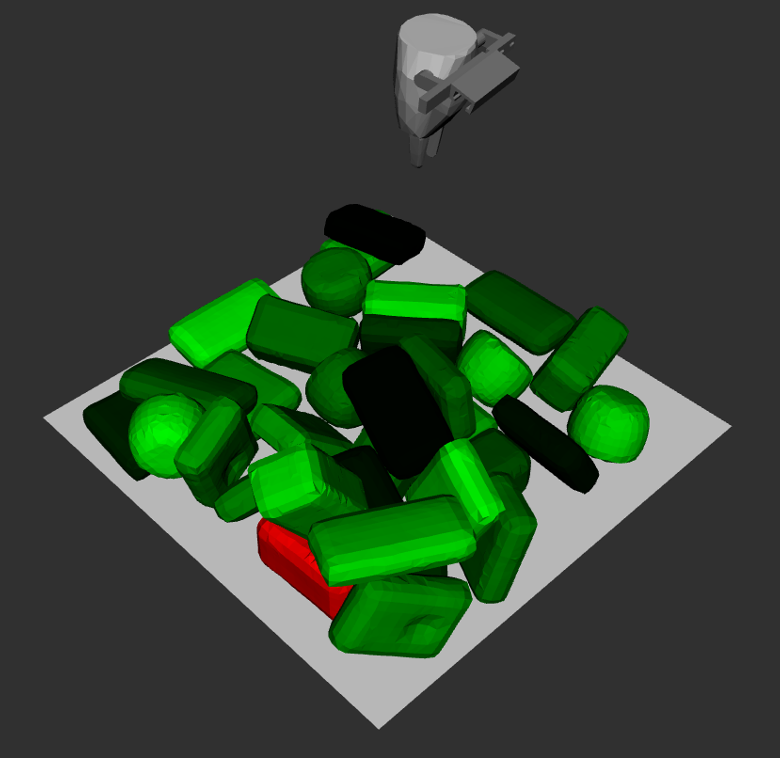}} &
    \multicolumn{2}{c}{\includegraphics[width=0.12\textwidth]{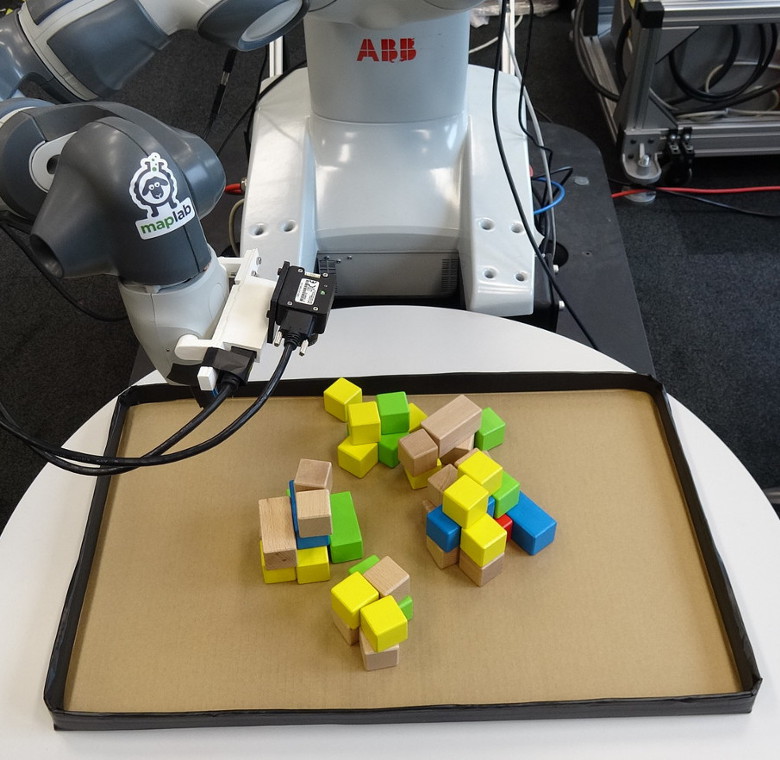}} \\ 
    &  \multicolumn{4}{c}{Cubes} &
    \multicolumn{2}{c}{Fixed size} & \multicolumn{2}{c}{Variable size} & \multicolumn{2}{c}{Random} & \multicolumn{2}{c}{Object} & \multicolumn{2}{c}{Real world} \\
    &  \multicolumn{2}{c}{Normal workspace} & \multicolumn{2}{c}{Large workspace} &
    \multicolumn{2}{c}{primitives} & \multicolumn{2}{c}{primitives} & \multicolumn{2}{c}{objects} & \multicolumn{2}{c}{models} & \multicolumn{2}{c}{shapes} \\ \midrule
    \multicolumn{1}{c}{GNBV} & $93.3$ & $8.59$ & $91.6$ & $31.89$ & $94.4$ & $8.48$ & $94.7$ & $8.26$ & $93.7$ & $8.62$ & $94.9$ & $13.25$ & $\boldsymbol{90.0}$ & $\boldsymbol{11.75}$ \\
    \multicolumn{1}{c}{A3DE} & $\boldsymbol{98.2}$ & $\boldsymbol{4.14}$ & $\boldsymbol{96.3}$ & $\boldsymbol{14.57}$ & $\boldsymbol{97.9}$ & $\boldsymbol{3.95}$ & $\boldsymbol{96.5}$ & $\boldsymbol{4.29}$ & $\boldsymbol{97.1}$ & $\boldsymbol{4.35}$ & $\boldsymbol{99.8}$ & $\boldsymbol{4.38}$ & $\boldsymbol{90.0}$ & $12.3$ \\ \midrule
    \multicolumn{1}{c}{GES} & $\boldsymbol{99.9}$ & $54.79$ & \multicolumn{2}{c}{/} & $\boldsymbol{99.6}$ & $51.72$ & $66.8$ & $70.84$ & $89.2$ & $54.42$ & $35.1$ & $93.15$ & $\boldsymbol{100.0}$ & $22.9$ \\
    \multicolumn{1}{c}{I3DE} & $88.0$ & $\boldsymbol{26.86}$ & \multicolumn{2}{c}{/} & $90.5$ & $\boldsymbol{21.22}$ & $\boldsymbol{85.0}$ & $\boldsymbol{30.08}$ & $\boldsymbol{89.5}$ & $\boldsymbol{20.82}$ & $\boldsymbol{82.1}$ & $\boldsymbol{29.79}$ & $\boldsymbol{100.0}$ & $\boldsymbol{15.35}$ \\
    \bottomrule 
    \end{tabular}
    }
    \caption{\revisedtext{Performance of the active and interactive agents trained on scenes with piles of \emph{cubes}, was evaluated with new scenes containing different objects. The numbers in each cell represent success rate (\%) and average number of steps per episode, respectively. Simulation results are obtained over $1000$ and real-world ones over $20$ episodes. Scenes \emph{fixed size primitives}, \emph{variable size primitives} and \emph{random shapes} contained more objects per pile ($15$ to $30$) in the interactive task in order to cover more the target object.}}
    \label{tab:scene_comparison}
\end{table*}

In simulation, we evaluated the performance of the agent on the active and interactive tasks.
Furthermore, we analyzed how well the agent generalizes to scenes that contain previously unobserved objects.
All the experiments are evaluated with $1000$ episodes. 
The simulation parameters for the active and interactive task are shown in Table~\ref{tab:experiment_parameters}.

\subsubsection{Active Perception}

\revisedtext{The agent for the active perception task was trained and evaluated using scenes with \emph{cubes} of fixed size.}
It learned to rise the camera to get an overview of the scene and to roll and yaw to observe the remaining unobserved parts.
\revisedtext{We evaluated the approach with a normal and a larger workspace.
The results are shown in Table~\ref{tab:scene_comparison} (Cubes) and Figure~\ref{fig:simulated_results_efficiency}.}
This efficient strategy shows a high success rate to solve the task ($98.2\%$ in normal scenes and $96.3\%$ in large ones). 
While our approach has larger success rate than the baseline, it is also more step-efficient, requiring roughly only half the amount of steps compared to the \acronym{gnbv}.
Additionally, the average time per step of the \acronym{a3de} is $8$ times smaller than the one of the \acronym{gnbv}.
\revisedtext{This is due to the sampling nature of the Next-Best-View algorithm and to the fact that the computation of the score of each action requires ray tracing, which is costly to obtain.
While this approach scales linearly with the number of samples, our approach operates in constant time as it consists in a single forward pass of the network.}
Finally, our agent achieved a $100\%$ success rate on scenes that did not contain a target object, meaning that it successfully learned when to terminate the episode.

\subsubsection{Interactive Perception}

\revisedtext{The agent for the interactive perception task was also trained and evaluated using scenes with piles of \emph{cubes} with fixed size.}
To complete the task, the agent learned to remain close to the surface and to move towards any clutter of objects sufficiently big to appear as unobserved in the \acronym{tsdf}.
The results of Figure~\ref{fig:simulated_results_efficiency} indicate that our learned approach for interactive perception is more efficient than the \acronym{ges} baseline requiring roughly half of the steps. 
Furthermore, the grid search does not scale well to an increase in the size of the workspace (quadratic in terms of the side length of the scene).
Our approach has a slightly lower success rate as shown in Table~\ref{tab:scene_comparison} (Cubes). 
The reason for this is that the \acronym{ges} baseline was specifically crafted for the \emph{cubes} scene, and therefore it achieves almost perfect success rate.
The baseline pre-computes all the steps the agent needs to take so time-wise it is comparable to ours.
Finally, our agent achieved a $96.1\%$ success rate on scenes without the target object meaning it successfully learned when to terminate an episode.

\subsubsection{Generalization}

\revisedtext{While the active and interactive agents were trained in a scene composed of \emph{cubes} of the same size, we evaluated them on four new scenes with different kinds of object piles: \emph{variable size primitives}, \emph{fixed size primitives}, \emph{random shapes}, and \emph{object models}. 
The results in Table~\ref{tab:scene_comparison} show that our method is able to generalize to new objects without fine-tuning and without significant loss of performance.
The \acronym{gnbv} also performs well for all the scenes, however, with lower success rate and larger number of steps than the \acronym{a3de}.
The \acronym{ges} significantly drops the success rate in scenes where the scale of objects changes.
}

\subsection{Real-World Results}

As the real experiment is a time-consuming and resource demanding task, the evaluation on the real setup for both active and interactive task is performed over $20$ runs. 
The scene is made of cubes and cuboids, with the parameters from Table~\ref{tab:experiment_parameters}.
The results are shown in Table~\ref{tab:scene_comparison}.

\subsubsection{Active Perception}

The comparison between the learned \acronym{a3de} and \acronym{gnbv} baseline shows comparable results to the ones obtained in simulation.
The average number of steps increased since the scenes we generated were more difficult than the ones in simulation (e.g. target hidden in a corner).
Nevertheless, this shows that the active task based on \acronym{rl} can be successfully transferred to reality with a high success rate.

\subsubsection{Interactive Perception}

The \acronym{i3de} method achieved a $100\%$ success rate, the same as the \acronym{ges} baseline. 
Furthermore, the learned interactive method outperforms the baseline in terms of average number of steps per episode. 
These results confirm the outcome of the evaluation in simulation and show that a transfer to reality can be performed without any fine-tuning on real data.

\subsection{Limitations}

One limitation of our system is that it is not aware of the size of the workspace and its current pose within it.
This can cause the agent to get stuck by trying to repeat the previous action, however not being able to execute it.

We trained our agent in simulation with a detached gripper, but when executing on the real robot, some end-effector poses are not kinematically feasible.
We overcame this limitation by restricting the workspace to feasible configurations only.

\revisedtext{Learning to terminate is challenging for the agent since it only has a compressed information about a local map, and if it is not big enough to cover the whole \acronym{tsdf} volume the agent has no way to know if it explored the whole area.}


\section{Conclusion}\label{sec:conclusion}

In this work, we have shown that an \acronym{rl} approach is feasible to learn a successful policy for an object finding task that requires both active and interactive perception.
By comparing our approach to two baselines, we have shown that it is more efficient than both the \acronym{gnbv} and the \acronym{ges}.
Our \acronym{a3de} agent was also more successful in every scene, while our \acronym{i3de} agent performs better on scenes with varying sized objects as it is able to better generalize to such cases.
Results obtained in the real robot experiments validate the findings provided by the simulations. 

In future work, we plan to integrate a more sophisticated object detection pipeline such that we do not rely on the fact that the target object has a specific color.
Furthermore, we plan to extend our volumetric representation to also encode dynamic objects and avoid the need to ``forget'' the volume where objects moved. 

\section*{Acknowledgements}
This work was supported by the Amazon Research Awards, the Swiss National Science Foundation (SNF) through the National Centre of Competence in Research (NCCR) on Digital Fabrication, the Luxembourg National Research Fund (FNR) 12571953, and ABB Corporate Research Center.


\bibliographystyle{IEEEtran}
\bibliography{references}

\end{document}